\documentclass[final]{cvpr}

\usepackage{times}
\usepackage{epsfig}
\usepackage{graphicx}
\usepackage{amsmath}
\usepackage{amssymb}
\usepackage{multirow}
\usepackage{booktabs}
\usepackage{pifont}
\usepackage{wrapfig}
\usepackage{graphbox}
\usepackage{makecell}

\usepackage[pagebackref=true,breaklinks=true,colorlinks,bookmarks=false]{hyperref}

\def\cvprPaperID{****} 
\def\confYear{CVPR 2021}
\newcommand\blfootnote[1]{%
  \begingroup
  \renewcommand\thefootnote{}\footnote{#1}%
  \addtocounter{footnote}{-1}%
  \endgroup
}

\begin{document}

\title{Exploring and Improving Mobile Level Vision Transformers}

\author{Pengguang Chen$^{1}$, Yixin Chen$^{1}$, Shu Liu$^{2}$, Mingchang Yang$^{1}$, Jiaya Jia$^{1,2}$  \\[0.2cm]
	The Chinese University of Hong Kong$^{1}$\quad SmartMore$^{2}$\\
	\{pgchen, yxchen, mcyang, leojia\}@cse.cuhk.edu.hk \quad liushuhust@gmail.com
}

\maketitle

\begin{abstract}
We study the vision transformer structure in the mobile level in this paper, and find a dramatic performance drop. We analyze the reason behind this phenomenon, and propose a novel irregular patch embedding module and adaptive patch fusion module to improve the performance.  We conjecture that the vision transformer blocks (which consist of multi-head attention and feed-forward network) are more suitable to handle high-level information than low-level features. The irregular patch embedding module extracts patches that contain rich high-level information with different receptive fields. The transformer blocks can obtain the most useful information from these irregular patches. Then the processed patches pass the adaptive patch merging module to get the final features for the classifier. With our proposed improvements, the traditional uniform vision transformer structure can achieve state-of-the-art results in mobile level. We improve the DeiT baseline by more than 9\% under the mobile-level settings and surpass other transformer architectures like Swin and CoaT by a large margin.
\end{abstract}
\blfootnote{Preprint.}
\section{Introduction}

Transformers were first proposed in the field of natural language processing (NLP), and have become the chosen one in NLP for a long time. Recently, ViT \cite{vit} introduced the traditional transformer architectures to computer vision tasks and achieved remarkable success, proving the potential of vision transformers towards computer vision. After that, many following works \cite{DeiT,swin,coat,localvit,contnet}, refine the vision transformer architectures to better fit vision tasks. The vision transformers can achieve comparable or even better performance than convolution neural networks (CNN).

Recent works mainly focus on improving the ultimate overall performance of vision transformers, which means they care more about boosting large vision transformers. However, the performance of vision transformers on the mobile level is still underexplored. In this paper, we investigate how existing vision transformer architectures perform in mobile level settings. As far as we know, this is the first work to comprehensively study current vision transformers in mobile level. We propose a novel irregular patch embedding module and adaptive patch merging module to improve the naive DeiT baseline to the state-of-the-art in mobile level vision transformers.

\begin{figure}
	\centering
	\includegraphics[width=\linewidth]{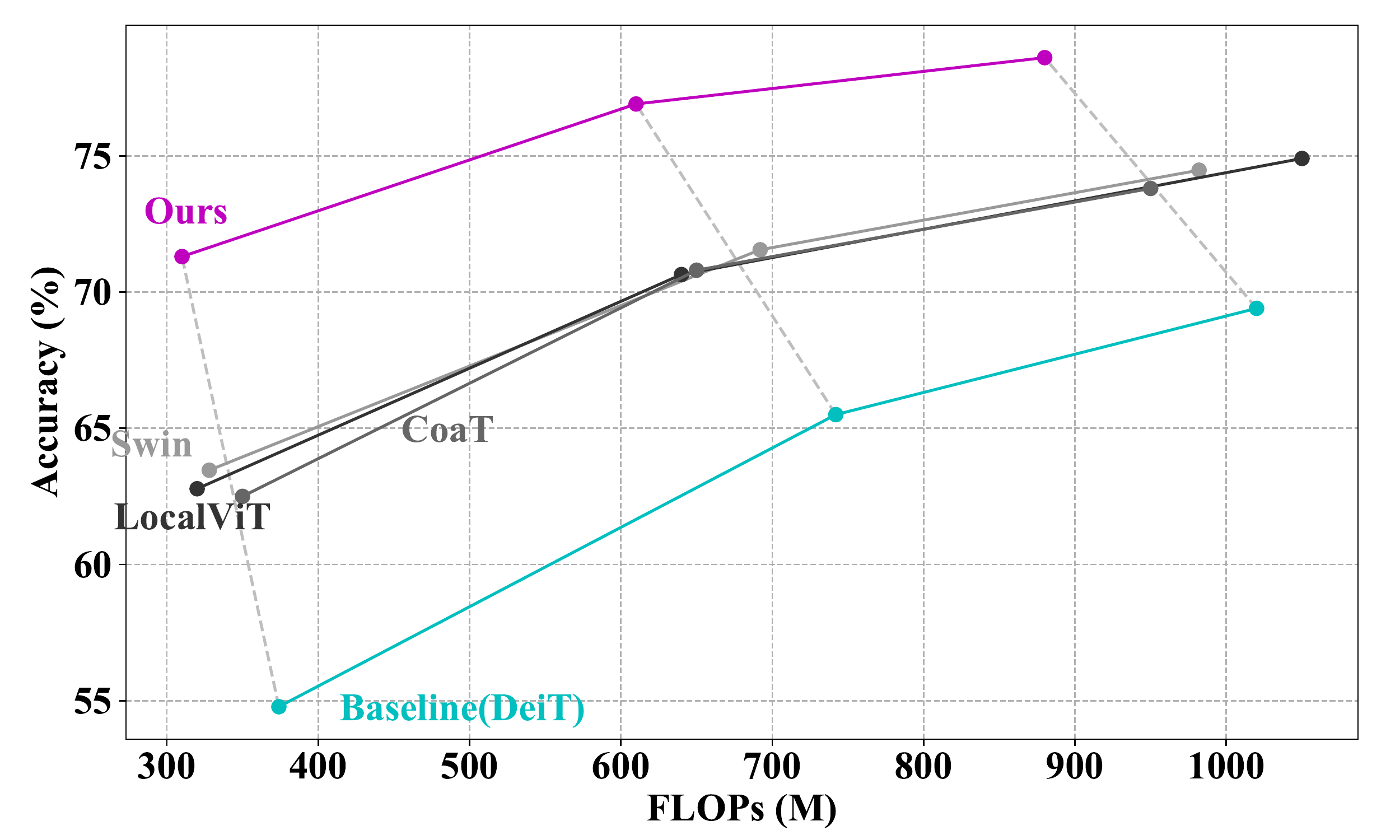}
	\caption{This figure compares the accuracy and FLOPs between our method and others on ImageNet. We propose two modules to improve the performance of traditional uniform vision transformers on the mobile level. As shown in this figure, it can improve the DeiT baseline by a large margin with fewer FLOPs. We can also surpass other recent vision transformers on the mobile level.}
	\label{fig:teaser}
\end{figure}

We find that, for existing vision transformers, once their FLOPs are compressed, the performance drop dramatically. For DeiT \cite{DeiT}, from DeiT-Base to DeiT-Small, the FLOPs are reduced by 1/4, the performance only declines by 2\%. Yet, from DeiT-Small to DeiT-Tiny, the accuracy lost by 7\%, though the FLOPs are also compressed by 1/4. When further compressing DeiT-Tiny to mobile level, the performance still experiences a noticeable decline. Other architectures \cite{swin,contnet} undergoing similar situations, we conjecture that these architectures are optimized towards large models which are equipped with the ability to extract features and avoid over-fitting, leading to low efficiency of information extraction. When the model is large enough, the drawbacks could be alleviated by enough parameters. However, when the size of models is under restriction, the performance can be highly affected.

Vision transformer blocks usually consist of two modules, multi-head self-attention (MSA) and feed-forward network (FFN). MSA module, which is designed to communicate between patches, has a strong ability to integrate high level information. But MSA fails to extract detailed low level information inside patches. Alternatively, FFN aims to extract per-patch information, but it only consists of fully-connected (FC) layers, which is inefficient to extract low level image features. When the vision transformer is big enough, the above limitations can be solved with large channels and deep structures. On the contrary, if the model is small, the performance will drop quickly. Many recent works \cite{localvit,contnet,levit} also notice this phenomenon, and they choose to perfect the MSA and FFN modules as well as adopt pyramid architectures like CNNs. However, we in this paper, show that the bottleneck may not lie in the MSA or FFN blocks, with slight improvements on the patch embedding module and class token, the original uniform structure of vision transformer blocks can achieve state-of-the-art results. 

Concretely, we find that the patch embedding of vision transformers behaves quite elementary, which only segments the input image into $14\times 14$ patches. The information inside one patch is simply transferred to the channel dimension. We argue that, for mobile-level models, this simple patch embedding leads to two aspects of problems. Firstly, the channel number is shrunken in small models. Directly project the information to channel dimensions will lose much of them. Secondly, the information in each patch is very primary, small models cannot learn well from such crude information. 

We propose three progressive improvements of the patch embedding for mobile level vision transformers. Firstly, we use a very simple module to generate patches from the input images, which is able to learn low level features from the original images and integrate them into patches. After that, the small-channel patches capture more high level information, which is suitable for transformer blocks. Secondly, we find that, with higher quality, the number of patches is not that important. The original DeiT has 14x14 input patches because every patch is primary, and adequate patches can preserve more information. But if we increase the quality, the number becomes unnecessary. So we decrease the number of patches to have more channels and blocks. Finally, we propose to use diverse patches to further improve the performance. Specifically, we generate different sizes of patches for transformer blocks. These diverse patches contain different levels of information, and transformer blocks have the great capability to handle them. 

Besides, we also notice that, there is a class token module in traditional uniform vision transformers. The class token is communicating with other patches and extracting information used for the final prediction. We discover that, in mobile level vision transformers, the number of blocks is small and the class token cannot extract adequate information, which highly affects the performance. We propose to replace the class token with a simple patch merging module to improve the accuracy on mobile level vision transformers.

We combine our proposed patch embedding and patch merging with DeiT and improve the performance by a large margin. More concretely, on 900M FLOPs models, we surpass DeiT by more than 9\%. On 600M and 300M FLOPs we improve DeiT by 11\% and 16\%, respectively. We also compare our method with other vision transformer architectures. Our method can surpass other models on the mobile level significantly.

\begin{figure*}[t]
	\centering
	\includegraphics[width=\linewidth]{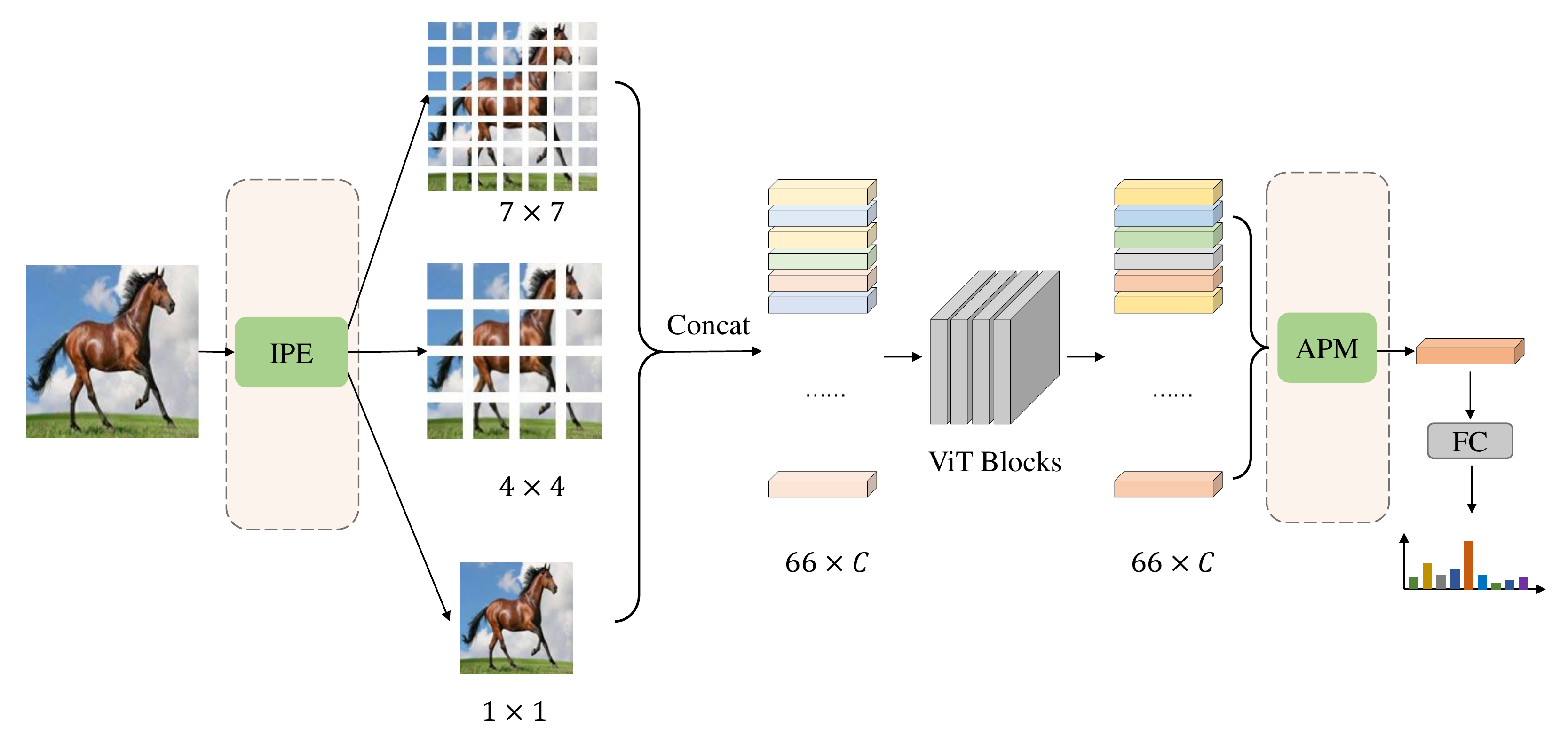}
	\caption{This figure shows the pipeline of our proposed modules. The modules in pink rounded rectangles are newly proposed by this paper. The images firstly go through the irregular patch embedding (IPE) module to generate $7\times7+4\times4+1\times1=66$ patches. These patches are concatenated together and being processed by uniform vision transformer blocks. Finally, the adaptive patch merging module (APM) gathers all patches to a single vector. The vectors are used by the fully-connected (FC) head for final prediction.}
	\label{fig:pipeline}
\end{figure*}

\section{Related Work}

\paragraph{Convolution neural networks.}
The rapid development of convolution neural networks (CNNs) originated from LeNet \cite{lenet}. While the past years have witnessed a series of convolution neural architectures \cite{alexnet,vgg,resnet,resnext} springing out to improve the limits, many researchers focus on designing efficient architectures to have better performance on budgeted computational resources. MobileNet \cite{mobilenet} proposed to use separable convolution to greatly reduce the FLOPs of CNNs. And many following works put forward skills to further improve the performance \cite{mobilenetv2,mobilenetv3}. 
Another series of ShuffleNet \cite{shufflenet,shufflenetv2} come up with channel shuffle operation to boost the performance on mobile level CNNs. Recently, many works \cite{mnas,fpnas,proxylessnas} use neural architecture search to find compact models with high performance. EfficientNet \cite{effnet,effnetv2} and RegNet \cite{regnet} design a family of networks with good trade-off between efficiency and performance. 

Another research focus is the use of self-attention in CNNs. Since \cite{attention} showed the great power of self-attention in the field of NLP, many works started to explore its usage in computer vision tasks. \cite{resatt} preliminarily introduces self-attention to CNNs. SENet \cite{senet} proposes a simple channel attention in CNNs and makes remarkable success. \cite{cbam,sknet,resnest,nonlocal} further explore self-attention's usage from the aspect of spatial, kernel, \etc.

\paragraph{Transformers.} Another important architecture,  transformers, was first introduced in \cite{attention} for machine translation. Transformers have now become the dominant architecture for NLP tasks, and a lot of works started to introduce the transformer architecture into computer vision tasks. \cite{imgtrans,sparsetrans} applied transformer on every pixel, which are primary and valuable attempts. ViT \cite{vit} firstly proposes the vision transformer architecture, which splits the image into patches and achieves comparable results with CNNs when pretrained on large datasets (JFT300M). The ViT architecture is very similar to the original transformers in NLP, which consists of a sequence of uniform self-attention blocks.
Further, DeiT \cite{DeiT} discover that the pretraining process is not necessary, and improves the training process of ViT. With aggressive data augmentations \cite{mixup,cutmix,randaug,repaug} and strong regularizations \cite{stodep,droppath}, DeiT can achieve comparable results with CNNs without large scale pretraining. Since then, the training cost of ViT is greatly reduced.

Many extraordinary works improve the original architecture of ViT. One typical promotion is to abandon the uniform architecture of ViT, and adopts the stage-wise like structure in CNNs \cite{vgg,resnet}. These works \cite{swin,contnet,coat,levit,pvt} propose to downsample the images progressively just like CNNs, and use transformer blocks on different resolutions to better suit the computer vision tasks. They think the stage-wise architecture can make use of the structure information in images and is more appropriate for vision tasks. Other researches keep the uniform architecture and improve the vision transformer from other aspects. Some of them \cite{localvit,tnt} improve the capability of transformer blocks by introducing more local information. Many works \cite{t2t,dynamicvit,earlyconv,dvt} leverage the patches to improve vision transformers.

However, these recent works mostly focus on boosting large vision transformers' performance. Although some of them improve the trade-off between accuracy and efficiency, the study on mobile-level vision transformers is still not yet explored. We, in this paper, investigate some famous vision transformers' performance on mobile-level, and show that with some simple improvements, the original uniform ViT structure can achieve state-of-the-art results on mobile-level vision transformers.


\section{Method}
We introduce our overall framework in this section. Firstly, we review the architecture of traditional vision transformers, and analyze the problems when compressing DeiT into mobile level. Secondly, we propose our novel patch embedding for mobile level vision transformers. Finally, we present a new method to replace the class token in mobile level vision transformers.

\subsection{Traditional Vision Transformer}
The most representative traditional vision transformer is DeiT, which consists of three components, patch embedding, transformer blocks and class token. We use DeiT to represent traditional vision transformer and analyze its components.

The first part of DeiT is the patch embedding module, which splits the input image into $14\times 14$ patches evenly. Each patch contains $16\times 16$ pixels, and it can be seen as a vector of size $16\times16\times3=768$. Then every patch is projected into a certain dimension using a fully-connected(FC) layer. These projected vectors are then forwarded as input to the following transformer blocks.

The main components are several transformer blocks. Each block consists of a multi-head self-attention (MSA) module and a fast-forward network (FFN) module. The MSA module adopts self-attention to realize the communication between patches, which captures the global information of the whole image. The FFN module is a multi-layer perceptron (MLP), which extracts the information inside every patch. We shall notice that, both the MSA module and FFN module are not experts at extracting local low-level information.

The final part is the class token. The class token is processed as a special patch, which is added to the $14\times14$ patches and these 197 patches will go through all transformer blocks together. The class token is regarded as an indicator to represent the class of the input image. After the transformer blocks, we pass the class token to a FC head to get the final prediction.

While the first and last part of DeiT is contributing little to the overall FLOPs, the majority of DeiT parameters are in the second part. When compressing the DeiT into the mobile level, it's a good plan to reduce the parameters of the second part. However, We show that, the existing patch embedding and class token are not well designed if we simply compress the transformer blocks. 

Following the official compression process from DeiT-Base to DeiT-Tiny, the compression of DeiT lies in two aspects, the channel of every patch and the number of transformer blocks. When we reduce the channel of patches, the information loss in patch embedding becomes very large. The channel number of DeiT-Base is 768, which equals the dimension of every patch in original images($16\times16\times3=768$). DeiT-Small and DeiT-Tiny reducing the channel number to 384 and 192, it will lose almost 1/2 and 3/4 information in the patch embedding blocks respectively. When we further compress the model to mobile-level, the information loss becomes quite severe and greatly affects the final accuracy. Therefore, a more advanced patch embedding module that can extract the information from the original image to a low-dimension patch is urgently needed.
What's more, when we reduce the number of transformer blocks, the drawbacks that MSA and FFN cannot extract low-level information effectively will be exposed. We can also use a carefully designed patch embedding module to overcome these drawbacks by extracting more high-level information into every patch. 

In addition, when we reduce the number of transformer blocks, the information communication between the class token and other patches is decreased to some extent. Thus, the information in the class token is not good enough for the final prediction, which also highly affects the accuracy. A more effective module to replace the class token is also essential for mobile level vision transformers.

\begin{figure}
	\centering
	\includegraphics[width=\linewidth]{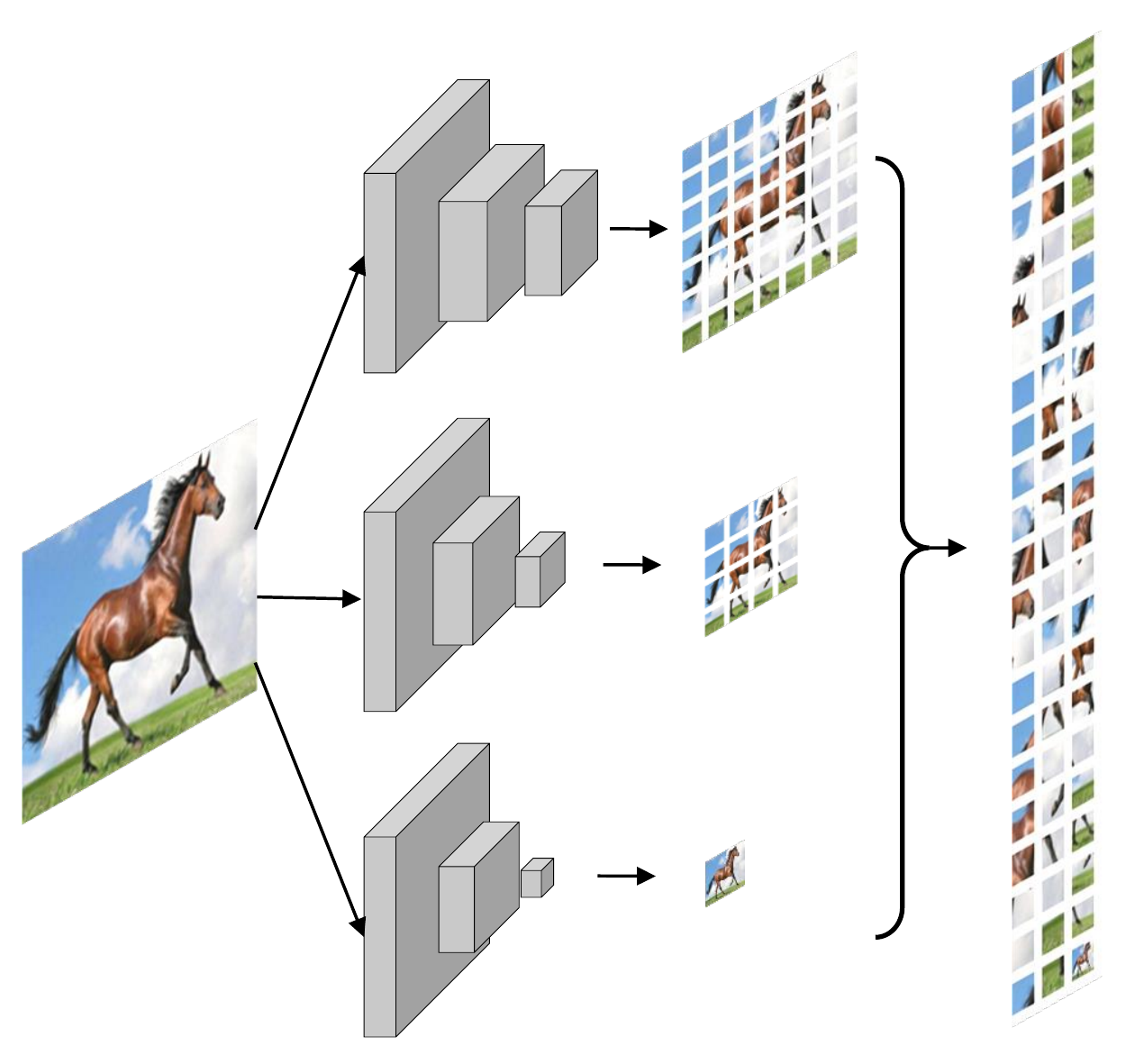}
	\caption{This figure shows the details of the irregular patch embedding module. The input image will go through three small CNNs with different receptive fields and generate patches with different levels' of information. Then these irregular patches are concatenated together to form the input for the following vision transformer blocks.}
	\label{fig:ipe}
\end{figure}

\begin{figure}
	\centering
	\includegraphics[width=0.9\linewidth]{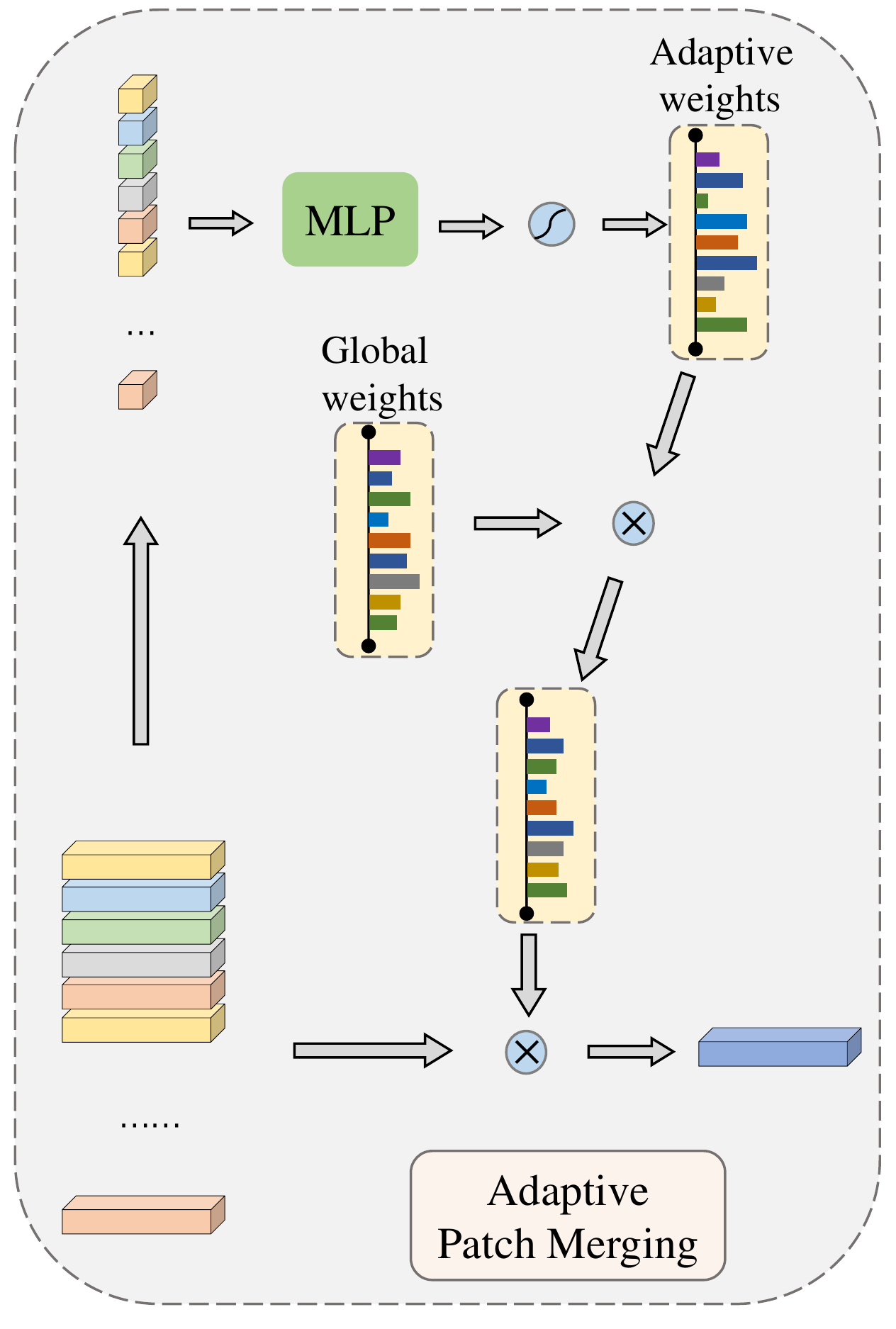}
	\caption{This figure shows the detailed architecture of our proposed adaptive patch merging module. There are two branches, one branch uses a multi-layer perceptron to generate adaptive weights for each image, the other is a global weight updated by gradient during training. We combine two weights together as the final weights for patches and merge them into one final patch.}
	\label{fig:apm}
\end{figure}

\subsection{Irregular Patch embedding}
\label{sec:ipe}
We propose a novel irregular patch embedding (IPE) module for mobile level vision transformers. We introduce our design in three steps progressively.

Firstly, we propose to use convolution layers in the patch embedding module. There are many advantages of using convolution layers. Many researches show that the convolution layer is skilled at extracting low-level local information, which is exactly a complement for transformer blocks. We can adopt a very elegant convolutional module to extract patches with rich high-level information for the following transformer blocks. Thus the information can be concentrated in low-dimensional patches, which effectively relieves the information loss in the traditional patch embedding. Therefore, using convolution layers as a patch embedding module is very suitable for mobile level vision transformers. More specifically, we use the combination of depth-wise convolution and point-wise convolution to further reduce resource consumption.

Secondly, we find that $14\times 14$ high-dimensional patches are too overwhelmed for mobile level vision transformers to afford. On the one hand, if we reduce the channels of patches and number of blocks to solve the above issue, the remaining transformer cannot sufficiently process that many patches. On the other hand, we are capable of reducing the number of patches while increasing the channels and numbers of blocks to find a better balance in the mobile level models. What's more, according to the first step, we can manage to generate quite high-quality patches, which contain more high-level information. The quality of patches helps relieve the influence of decreased patches. To some extent, the quality of patches is even more crucial than the number of patches in mobile level vision transformers.

Thirdly, we produce irregular patches in mobile level vision transformers. DeiT splits the input image into $14\times14$ parts evenly and obtains regular patches from them, making these patches have the same-level receptive field. However, the type of classes in ImageNet dataset varies and these classes exhibit quite dissimilar characteristics. Some object classes like dog and cat are more likely to be discriminated based on small-region high-level information, while the whole-image texture-level information turns out to be more important for other scene classes like cliff and lakeside. Therefore, using regular patches is not efficient. We tend to generate irregular patches which contain different receptive fields to further improve mobile level vision transformers' performance. The detailed architecture is shown in Figure \ref{fig:ipe}. Concretely, instead of $8\times 8$ patches, we use $7\times7+4\times4+1\times1$ patches. These patches are generated with three parallel modules which consist of inverted residual bottleneck blocks with SE layer. We use different numbers of strides to generate patches with different receptive fields.
The vision transformer blocks own the great power to effectively communicate between these patches and excavate the most useful information for final prediction.

\subsection{Adaptive Patch Fusion}
In DeiT, there is a learnable class token to predict the classification results. The class token is randomly initialized and optimized via gradient descent during training. The class token is concatenated with patches together to be processed by vision transformer blocks. In order to extract useful information for recognition, the class token communicates with other patches in the MSA module. When it comes to the final prediction, only the class token is used, and the patches are abandoned. This design is suitable for large models, where the class token naturally gathers enough information from adequate transformer blocks. In that case, other patches' information is not necessary, and the participation of other patches may even harm the prediction. 

However, the situation in the mobile level vision transformer is very different. We have to carefully utilize every patches' information under small computation limitations. We find that adopting the class token is much worse than integrating all patches' information in mobile level vision transformers. We conjecture that the class token cannot gather enough information with a limited number of transformer blocks. Therefore, how to effectively using all patches' information is a key problem in mobile level vision transformers.

\begin{table}[t]
    \centering
    \begin{tabular}{c|c c c c}
        \toprule
         FLOPs & Channel & Depth & \# Heads & MLP Ratio   \\
         \midrule
         880M & 300 & 8 & 12 & 4 \\
         610M & 264 & 6 & 12 & 4 \\
         310M & 210 & 5 & 10 & 4\\
         \bottomrule
    \end{tabular}
    \vspace{0.1in}
    \caption{Detailed architectures of different size models.}
    \label{tab:arch}
\end{table}

\begin{table*}
	\centering
	\begin{tabular} {l | l | l | c | c | c}
		\toprule
		Group & Method & FLOPs & \makecell[c]{ImageNet\cite{imagenet} \\ Top-1  (\%)} & \makecell[c]{Real\cite{real}\\ Top-1  (\%)} & \makecell[c]{V2 \cite{imagenetv2} \\ Top-1  (\%)}\\
		\midrule
		\multirow{5}{*}{CNNs} & ResNet18 \cite{resnet} & 1.8G & 69.8 & 77.3 & 57.1\\
		& ResNet50 \cite{resnet} & 4.2G & 76.2 & 82.5 & 63.3\\
		& MobileNet \cite{mobilenet} & 575M & 70.6 & - & - \\ 
		& MobileNetV2 \cite{mobilenetv2} & 300M & 73.0 & 82.1 & 60.2 \\
		& MobileNetV3 \cite{mobilenetv3} & 219M & 75.2 & 82.2 & 63.4 \\
		& EfficientNet B0 \cite{effnet} & 390M & 77.1 & 83.5 & 64.3 \\
		& EfficientNet B1 \cite{effnet} & 700M & 79.1 & 84.9 & 66.9 \\
		& EfficientNet B2 \cite{effnet} & 1000M & 80.1 & 85.9 & 68.8 \\
		
		\midrule
		\midrule
		\multirow{5}{*}{900M FLOPs and More} & DeiT \cite{DeiT} & 1020M & 69.4 & 77.4 & 56.8 \\
		& Swin \cite{swin} & 980M & 74.5 & 81.8 & 62.4 \\
		& LocalViT \cite{localvit} & 1050M & 74.9 & 82.2 & 62.9 \\
		& CoaT \cite{coat} & 950M & 73.8 & 81.4 & 62.1\\
		& Ours & 880M & \textbf{78.6} & 84.5 & 66.1 \\
		\midrule
		\multirow{5}{*}{Around 600M FLOPs} & DeiT \cite{DeiT} & 740M & 65.8 & 73.5 & 53.5 \\
		& Swin \cite{swin} & 690M & 71.6 & 79.4 & 59.0 \\
		& LocalViT \cite{localvit} & 640M & 70.6 & 78.7 & 59.1 \\
		& CoaT \cite{coat} & 650M & 70.8 & 78.6 & 58.0 \\
		& Ours & 610M & \textbf{76.9} & 83.6 & 64.4 \\
		\midrule
		\multirow{5}{*}{Around 300M FLOPs} & DeiT \cite{DeiT} & 370M & 54.8 & 62.3 & 43.2 \\
		& Swin \cite{swin} & 330M & 63.5 & 71.5 & 51.0\\
		& LocalViT \cite{localvit} & 320M & 62.8 & 70.8 & 50.8 \\
		& CoaT \cite{coat} & 350M & 62.5 & 70.6 & 50.7 \\
		& Ours & 310M & \textbf{71.3} & 79.1 & 58.7 \\
		\bottomrule
	\end{tabular}
	\label{tab:imagenet}
	\vspace{0.1in}
	\caption{We split models into different groups for better comparison. The first group "CNNs" summarizes SOTA convolution networks' results on ImageNet. Other groups are vision transformers with different FLOPs. Other transformers' results are reproduced under authors' released official codes with constrained FLOPs.}
\end{table*}

To tackle the above problem, we replace the class token with a designed simple patch merging module that can efficiently gather all patches' information. In Section \ref{sec:ipe}, we have mentioned that we are using irregular patches, and different images may focus on different patches. Based on this, we propose to adaptively fuse patches, which means we assign the patches different weights according to each image. Then we gather all patches together. The adaptive patch fusion module has a quite simple mechanism as shown in Figure \ref{fig:apm}. We generate an adaptive weight for each image and maintain a global weight for all images. The final weights is obtained by multiplying these two weights together. The adaptive weights are calculated using a multi-layer perceptron. The global weights are updated by gradients during training and fixed during testing, which is computed by a linear layer. This design is much more advantageous than the class token to mobile level vision transformers. More experimental results and visualized analyses are available in Section \ref{sec:anapm}.

Overall, we propose IPE and APM to boost the performance of traditional uniform vision transformers like DeiT. We set up different sizes of transformers, the details are summarized in Figure \ref{tab:arch}.


\section{Experiments}

\subsection{ImageNet Classification}

\paragraph{Dataset} ImageNet \cite{imagenet} is the most challenging dataset for image classification task. This dataset spans 1000 classes with around 1.3 million images. Each class has around 1300 training images and 50 validation images. 

\paragraph{Training Details}
We follow the default training setting from \cite{DeiT} except for batchsize and learning rate. Concretely, we train our models on four 2080Ti GPUs for 300 epochs, and the batchsize for each GPU is 128. Following \cite{onehour}, we linearly rescale our learning rate to 0.0005. The weight decay is set to 0.05. We use AdamW optimizer with the epsilon being 1e-8. We use strong data augmentation following \cite{DeiT}, including Mixup \cite{mixup}, CutMix \cite{cutmix}, AutoAug \cite{autoaug}, color jitter, Random Erase \cite{re}, \etc. We also adopt label smoothing and DropPath just like \cite{DeiT}.

\begin{table*}
	\centering
	\begin{tabular}{c c c c c c c}
		\toprule 
		Model & FLOPs & Pretrain & CIFAR-10  & CIFAR-100  & iNatural-18 & iNatural-19 \\
		\midrule
		DeiT & 1020M &  & 85.2 & 65.3 & 53.1 & 61.7\\
		DeiT & 1020M & \ding{52} & 97.9 & 85.0 & 62.1 & 71.2 \\
		Ours & 880M & & 93.6 & 77.8 & 56.0 & 70.4\\
		Ours & 880M & \ding{52} & 98.1 & 85.3 & 64.3 & 71.6\\
		\midrule
		DeiT & 740M &  & 84.1 & 62.4 & 47.9 & 57.6\\
		DeiT & 740M & \ding{52} & 97.5 & 84.4 & 58.0 & 69.6\\
		Ours & 610M & & 93.5 & 78.4 & 53.2 & 70.9\\
		Ours & 610M & \ding{52} & 97.7 & 84.9 & 62.8 & 71.4 \\
		\midrule
		DeiT & 370M &  & 80.2 & 56.2 & 36.2 & 48.0 \\
		DeiT & 370M & \ding{52} & 96.8 & 81.1 & 46.0 & 60.6\\
		Ours & 310M & & 92.0 & 77.2 & 51.4 & 64.3\\
		Ours & 310M & \ding{52} & 97.4 & 83.7 & 59.3 & 68.2 \\
		\bottomrule
	\end{tabular}
	\label{tab:cifar}
	\vspace{0.1in}
	\caption{This table shows the top-1 accuracy on the CIFAR-100 classification task for different models with or without pretraining.}
\end{table*}

\paragraph{Results}
The results are summarized in Table \ref{tab:imagenet}. We can easily conclude that, our method surpasses previous vision transformer models by a large margin on the mobile level. According to models' FLOPs, we separate them into three different groups: 900M FLOPs and more, around 600M FLOPs, and around 300M FLOPs. In the group of 900M FLOPs, we surpass our baseline method DeiT \cite{DeiT} by 9\%. And we beat the previous state-of-the-art (SoTA) method by around 2.5\% with fewer FLOPs. When we compress all methods to the group of 600M FLOPs, our accuracy is 11\% higher than DeiT and 5\% higher than the previous SoTA. Finally, in the 300M FLOPs group, our method can further outperform DeiT by 16\%, and our result is around 7\% better than the previous SoTA. The different groups' results manifest our method's superiority against others especially when the models are slimmer. Plus, these results verify the effectiveness of our proposed modules. With a very strict computation constraint, our proposed irregular patch embedding module is able to efficiently extract low-level informative image patches. Also, the adaptive patch merging module could make very thorough use of every patch's information. Combining these modules together, ours makes a remarkable difference even with the most original vision transformer blocks.

\subsection{Downstream Tasks}

We also evaluate our proposed method on other datasets. CIFAR-10 and CIFAR-100 are relatively small datasets that contain 50,000 images for training and 10,000 images for validation. The size of images in the CIFAR datasets is only $32\times 32$. We upsample them to $224\times 224$ following DeiT \cite{DeiT}. But the number of training epochs in DeiT is too large (7200 epochs), we use extremely fewer epochs (300 epochs) in order to save tremendous training costs. Besides, we evaluate our method on iNautural-18 and iNatural-19 datasets. We also resize the images to $224\times 224$ and train models for 300 epochs. The result is summarized in Table \ref{tab:cifar}. We can find that, our methods' performance is much better than the baseline DeiT method on downstream tasks.

\begin{table*}[t]
	\centering
	\begin{tabular}{c | c c c | c}
		\toprule
		ID & Use Conv in PE & Reduce Patch Number & Irregular Patch embedding & Top-1 Accuracy (\%) \\
		\midrule
		(a) & & & & 65.79 \\
		\midrule
		(b) & \ding{51} & & & 69.90 \\
		\midrule
		(c) & \ding{51} & \ding{51} & & 72.65 \\
		\midrule
		(d) & \ding{51} & \ding{51} & \ding{51} & \textbf{73.68} \\
		\bottomrule
	\end{tabular}
	\label{tab:pe}
	\vspace{0.1in}
	\caption{This table shows our 900M model's results on ImageNet with different patch embedding methods. Models in this table are trained for 100 epochs to save resources.}
\end{table*}

\section{Analysis}

We analyze several different components in our proposed framework in this section. Firstly, we show the improvements brought by the IPE module in Section \ref{sec:anape}. Then, we show the effectiveness of APM and visualize different images patch weights in Section \ref{sec:anapm}. Finally, we analyze the influence of positional encoding in our framework in Section \ref{sec:anapos}

\subsection{Patch embedding}
\label{sec:anape}

We experiment with different types of patch embedding and summarize the results in Table \ref{tab:pe}. We report different models' top-1 accuracy on ImageNet. We only train these models for 100 epochs to save resources. Please note that, we only change the patch embedding in this table. (The channel of patches and number of transformer blocks is changed in order to meet the FLOPs constraint.) And we have the class token kept and do not adopt our proposed adaptive patch merging module.

The model (a) is the DeiT-tiny baseline whose FLOPs is around 1200M. The model (b)'s FLOPs is also around 1200M. Other models' are around 900M. As is shown in this table, when we introduce convolution layers to the patch embedding (PE) module, the performance is significantly improved. This is because the convolution layers are good at processing low-level features and extracting more useful information into the low-dimensional features for every patch. Comparing (b) and (c), the performance is improved with fewer FLOPs. This phenomenon verifies our hypothesis that the quality of patches is more critical than the quantity. Fewer high-quality patches is more practical than a large amount of low-dimensional patches. After we split the patches into different irregular patches, the performance is further improved as shown in (d).

\subsection{Patch Merging}
\label{sec:anapm}
\begin{table}
	\centering
	\begin{tabular}{c c c}
		\toprule
		ID & Method & Top-1 Accuracy (\%) \\
		\midrule
		(a) & Class Token & 73.68 \\
		(b) & Average Pooling & 74.83 \\
		(c) & Adaptive Patch Merging & \textbf{75.59} \\
		\bottomrule
	\end{tabular}
	\label{tab:pm}
	\vspace{0.1in}
	\caption{This table shows our 900M model's results on ImageNet with different final feature generation methods. Models in this table are trained for 100 epochs to save resources.}
\end{table}

\newcommand{\figsizea}{0.2\linewidth}
\newcommand{\figsizeb}{0.222\linewidth}
\begin{figure*}[t]
	\centering
	\resizebox{0.9\linewidth}{!}{
		\begin{tabular}{cccccc}
			Input Image & Patch Weight & Input Image & Patch Weight & Input Image & Patch Weight \\
			\includegraphics[align=c,width=\figsizea]{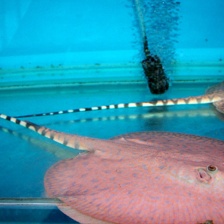} &
			\includegraphics[align=c,width=\figsizeb]{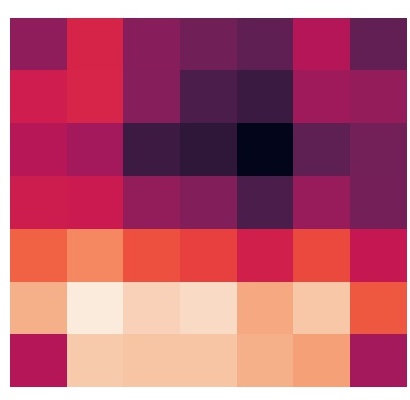} &
			\includegraphics[align=c,width=\figsizea]{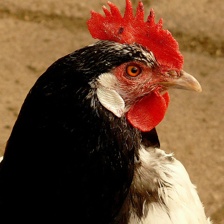} &
			\includegraphics[align=c,width=\figsizeb]{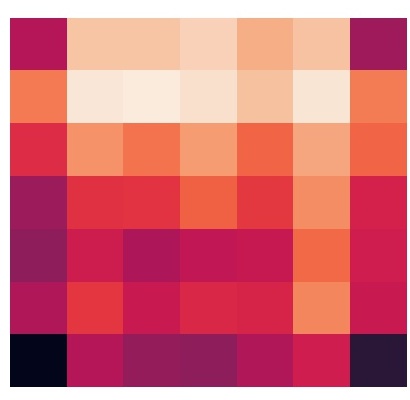} &
			\includegraphics[align=c,width=\figsizea]{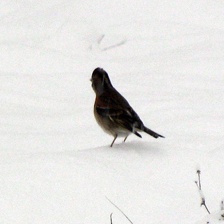} &
			\includegraphics[align=c,width=\figsizeb]{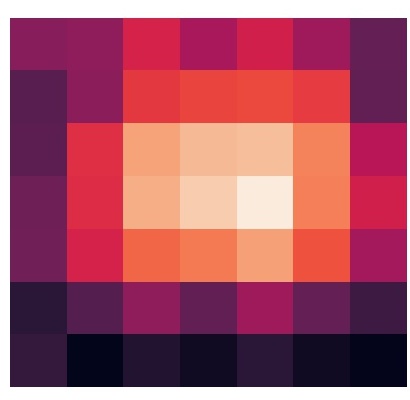} \\
			\includegraphics[align=c,width=\figsizea]{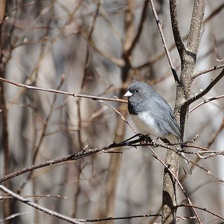} &
			\includegraphics[align=c,width=\figsizeb]{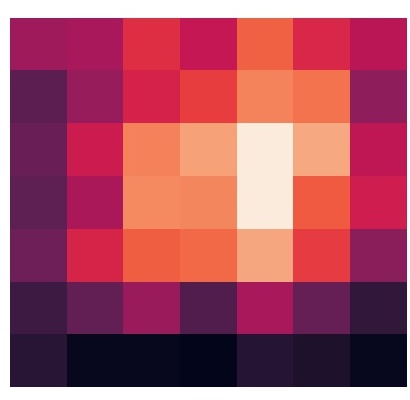} &
			\includegraphics[align=c,width=\figsizea]{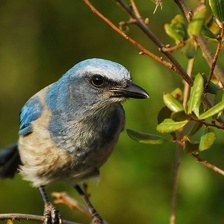} &
			\includegraphics[align=c,width=\figsizeb]{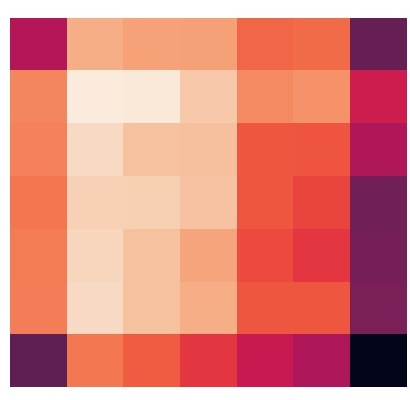} &
			\includegraphics[align=c,width=\figsizea]{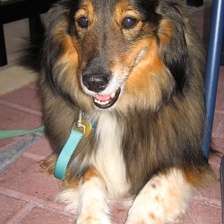} &
			\includegraphics[align=c,width=\figsizeb]{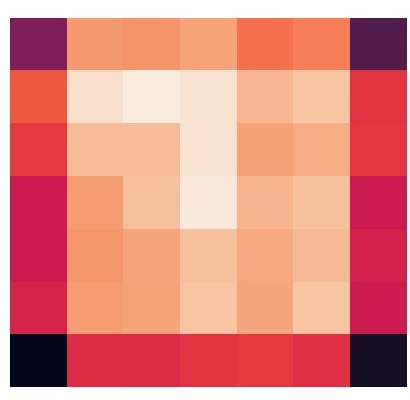} \\
		\end{tabular}}
	\caption{This figure visualizes the weights of $7\times 7$ patches in APM. Lighter color means larger value.}
	\label{fig:pm}
\end{figure*}

We experiment with different patch merging methods in this section. We also train the models on ImageNet for 100 epochs and report their top-1 accuracy. In Table \ref{tab:pm}, we only manipulate the patch merging method, and we always use the irregular patch embedding. 

Comparing (a) and (b), we surprisingly find that the simplest average pooling is much better than the class token, which is contrary to other large vision transformers. We conjecture in mobile level vision transformers, the class token cannot effectively gather adequate information with fewer transformer blocks. And the average pooling can also integrate all patches together to achieve better performance. 

Based on the observation of (b), we propose our adaptive patch merging module, which can gather patches information effectively. As shown in Table \ref{tab:pm} (c), our proposed module can further improve the performance of average pooling.

We also visualize some images' patch merging weights in Figure \ref{fig:pm}. The input image and patch weight for the input are listed. The lighter means a larger value in the patch weight. We can easily conclude that for small objects like birds, the patch weight will focus on the target objects' area. For objects with obvious characteristics like a rooster, the patch weight can correctly focus on the most discriminative parts. Please note that, for most images, the patch weights of corners are relatively small, which is consistent with our intuition that corners are not important to classify these images.

\subsection{Positional Encoding}
\label{sec:anapos}

We also investigate the usage of positional encoding in our framework. Since we didn't change the transformer blocks of DeiT, we continue to use the same positional encoding as DeiT. And we show the results of our model with and without positional encoding in Table \ref{tab:pos}. We find that, the positional encoding is not essential for our model. The results in Table \ref{tab:pos} are quite similar. We conjecture that, the irregular patch embedding module already encodes enough positional information inside the patch, and this kind of positional encoding doesn't provide extra valuable information for the transformer blocks. A more suitable positional encoding may further improve the performance, we leave this part for future study.

\begin{table}[h]
	\centering
	\begin{tabular}{c @{\hspace{0.5in}}c}
		\toprule
		Method & Top-1 Accuracy (\%)\\
		\midrule
		w/ positional encoding & 75.59 \\
		w/o positional encoding & 75.52 \\
		\bottomrule
	\end{tabular}
	\label{tab:pos}
	\vspace{0.1in}
	\caption{This table compares our models' performance with and without positional encoding.}
\end{table}

\section{Conclusion}

In this paper, we analyze the drawbacks of traditional vision transformer blocks in mobile level models. We compress some typical vision transformer architectures into mobile level, and find a dramatic performance drop. We propose an irregular patch embedding module and an adaptive patch merging module to upgrade the original uniform vision transformer structures in mobile level. Our improved vision transformer surpasses the baseline model by a large margin, and outperforms state-of-the-art transformers significantly. 

We show the traditional uniform vision transformer structure can achieve state-of-the-art results with some refinements. Improving the structure may further boost the performance. And a more suitable positional encoding is also worth studying.

{\small
\bibliographystyle{ieee_fullname}
\bibliography{egbib}
}

\end{document}